\documentclass[letterpaper, 10 pt, conference]{ieeeconf}  

\IEEEoverridecommandlockouts                              

\usepackage{graphicx}
\usepackage{tikz}
\usepackage{comment}
\usepackage{amsmath,amssymb} 
\usepackage{tabularx}
\usepackage{multirow}

\usepackage{subfig}
\usepackage{booktabs}

\usepackage{threeparttable}
\usepackage{adjustbox}

\usepackage{eqparbox}
\usepackage{arydshln}
\usepackage{wrapfig}
\usepackage{threeparttable}
\usepackage{adjustbox}
\usepackage{eqparbox}
\usepackage{arydshln}
\usepackage{wrapfig}
\usepackage{interval}

\usepackage{algorithm}

\usepackage{amsmath,algorithm,tabularx}
\usepackage[noend]{algpseudocode}
\usepackage{hyperref}

\makeatletter
\newcommand{\multiline}[1]{%
  \begin{tabularx}{\dimexpr\linewidth-\ALG@thistlm}[t]{@{}X@{}}
    #1
  \end{tabularx}
}
\makeatother

\newcommand{\distas}[1]{\mathbin{\overset{#1}{\kern\z@\sim}}}%
\newsavebox{\mybox}\newsavebox{\mysim}
\newcommand{\distras}[1]{%
	\savebox{\mybox}{\hbox{\kern3pt$\scriptstyle#1$\kern3pt}}%
	\savebox{\mysim}{\hbox{$\sim$}}%
	\mathbin{\overset{#1}{\kern\z@\resizebox{\wd\mybox}{\ht\mysim}{$\sim$}}}%
}

\title{Improving LiDAR 3D Object Detection via Range-based Point Cloud Density Optimization}

\author{Eduardo R. Corral-Soto$^{1}$, Alaap Grandhi$^{1}$, Yannis Y. He$^{1}$, Mrigank Rochan$^{1,2}$ and Bingbing Liu$^{1}$ 
	\thanks{$^{1}$All authors are with Huawei Noah's Ark Lab, Canada at the time of writing.  
		{\tt\small \{eduardo.corral.soto, alaap.grandhi, yannis.yiming.he, liu.bingbing  \}@huawei.com}}%
	\thanks{$^{2}$Mrigank Rochan is with the Department of Computer Science University of Saskatchewan.  
		{\tt\small mrochan@cs.usask.ca}}%
}

\begin{document}

\maketitle
\thispagestyle{empty}
\pagestyle{empty}

\begin{abstract}

In recent years, much progress has been made in LiDAR-based 3D object detection mainly due to advances in detector architecture designs and availability of large-scale LiDAR datasets. Existing 3D object detectors tend to perform well on the point cloud regions closer to the LiDAR sensor as opposed to on regions that are farther away. In this paper, we investigate this problem from the data perspective instead of detector architecture design. We observe that there is a learning bias in detection models towards the dense objects near the sensor and show that the detection performance can be improved by simply manipulating the input point cloud density at different distance ranges without modifying the detector architecture and without data augmentation. We propose a model-free point cloud density adjustment pre-processing mechanism that uses iterative MCMC optimization to estimate optimal parameters for altering the point density at different distance ranges. We conduct experiments using four state-of-the-art LiDAR 3D object detectors on two public LiDAR datasets, namely Waymo and ONCE. Our results demonstrate that our range-based point cloud density manipulation technique can improve the performance of the existing detectors, which in turn could potentially inspire future detector designs.

\end{abstract}

\section{Introduction}
\label{sec:intro}

The performance of LiDAR 3D object detectors for autonomous driving has improved significantly in the recent years, thanks mainly to the evolution of deep learning detection architectures, and to the emergence of public labeled LiDAR point cloud datasets such as Waymo ~\cite{sun2020scalability}, NuScenes ~\cite{caesar2020nuscenes}, KITTI ~\cite{geiger2013vision}, and ONCE ~\cite{mao2021one}. The evolution of such detection architectures includes methods that voxelize the point cloud and either employ 3D convolutions ~\cite{zhou2018voxelnet}, or 2D convolutions ~\cite{lang2019pointpillars}, ~\cite{yin2021center}, methods that operate on projections of LiDAR points ~\cite{chen2017multi}, ~\cite{ku2018joint}, ~\cite{corral2020understanding}, methods that operate directly on 3D points ~\cite{qi2017pointnet++}, and more recently, methods that combine voxelization with point-level processing ~\cite{shi2020pv}. For a comprehensive chronological overview of 3D object detection algorithms using LiDAR data see ~\cite{wu2020deep} .

\begin{figure} 
	\centering	
		\includegraphics[width=0.8\columnwidth, trim={0cm 13.5cm 0cm 0cm},clip]{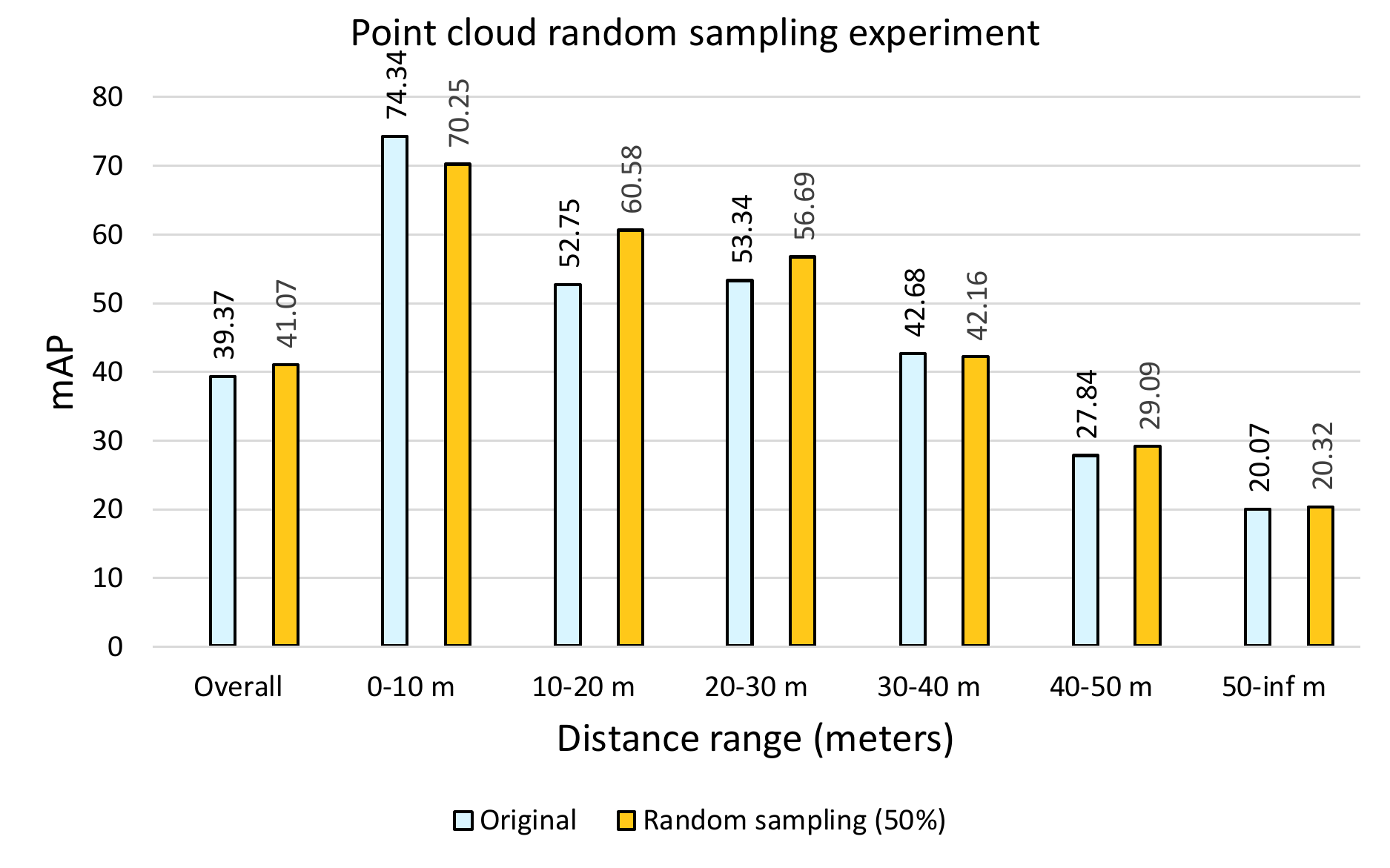} 		
	\caption{mAP for Centerpoint 3D object detector ~\cite{yin2021center} trained/evaluated on Waymo point clouds with and without random sampling with $50\%$ keep percentage on distance range [0-10] m. We computed mAP separately for different distance ranges, and observed that performace improved at farther away ranges, which suggests the existence of a learning bias toward the closer denser objects.  
	\label{Motivation_random_sampling_bar_plots} }
\end{figure}

Data augmentation is a set of techiques to improve performance without detector architectural modifications by introducing diversity in the training data. Typical augmentation techniques include: 1) global perturbations such as scaling, translation, rotation of the point cloud, 2) local perturbations of objects ~\cite{yang20203dssd}, and 3) insertion of database objects from the same or different datasets ~\cite{fang2021lidar}, ~\cite{ren2022object}. See ~\cite{hahner2020quantifying} for a taxonomy of data augmentation techniques in LiDAR 3D object detection.

Domain adaptation methods ~\cite{wang2020train}, ~\cite{yang2021st3d}, ~\cite{corral2021lidar} have been used as a form of data augmentation by shifting the distribution of a large labeled source training dataset into that of a small target dataset to increase the amount of training data. Recent GAN-based object-level shape completion methods ~\cite{tsai2022viewer}, ~\cite{chen2019unpaired}, ~\cite{zhang2021unsupervised} have also been used to either complete occluded point clouds, or manipulate their distribution to attempt improving performance. However, these methods have mainly focused on synthetic data, with very limited evaluations on real LiDAR object-level (cars) point clouds.

\begin{figure} 
	\centering	
		\includegraphics[width=0.8\columnwidth, trim={0cm 9cm 0cm 0cm},clip]{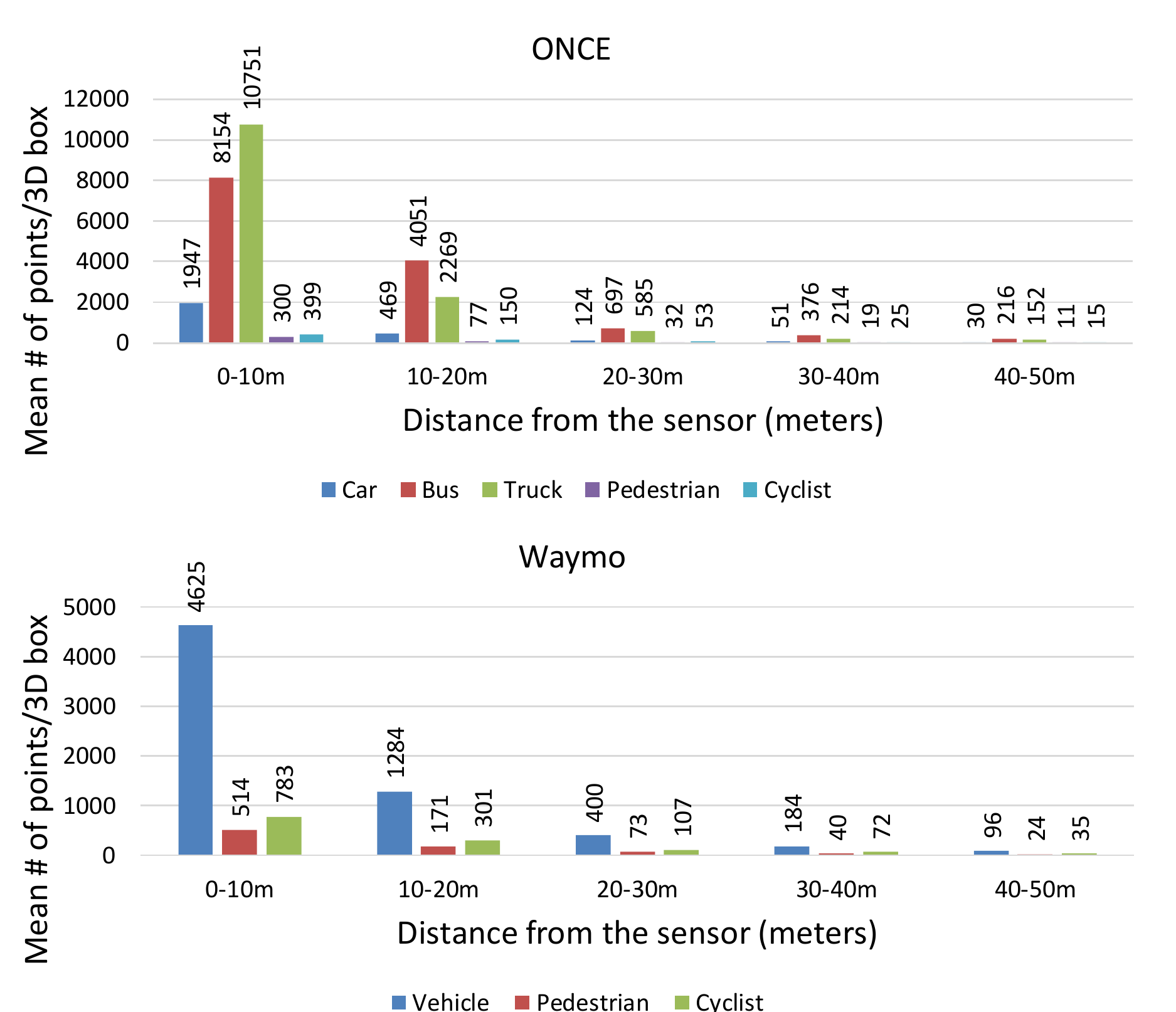} 		
	\caption{Mean number of object-level points in a 3D bounding box for different Euclidean distance ranges from the LiDAR sensor. (top) ONCE, (bottom) Waymo.
	\label{Mean_density_bar_plots_2} }
\end{figure}

\textbf{Problem}. It is well known that 3D point density distribution accross distance ranges of LiDAR point clouds for autonomous driving have a highly non-uniform distribution ~\cite{hu2022point}, ~\cite{corral2020understanding}. This non-uniformity is mainly caused by the fixed scanning pattern of the LiDAR sensor and by its limited beam resolutions, which results in large density discrepancies between objects located at different distance ranges from the sensor. For example, in Fig. \ref{Mean_density_bar_plots_2} cars from different datasets, located between $0$ and $10$ meters from the sensor contain between around $2000$ and $5000$ 3D points, compared to cars located between $40$ to $50$ meters which only contain between $30$ to $100$ points. In Fig. \ref{Mean_density_bar_plots_2} each bar represents the average number of points that fall inside of a 3D bounding box for each class at different Euclidean distances (meters) from the LiDAR sensor. The density-adaptive sampling method from ~\cite{arief2019density} addresses the non-uniform density problem by voxelizing point clouds and applying over-sampling to balance the per-voxel density based on an average density, followed by an empirical sampling technique. More recently, the Point Density-Aware Voxel network 3D object detection method (PDV) from ~\cite{hu2022point} addresses this problem architecturally by taking into account point density variations. The method localizes voxel features through voxel point centroids to spatially localize features and aggregate them through a KDE-based density-aware RoI grid pooling module and self-attention. 

\textbf{Our contributions}. Instead of focusing on developing a new 3D object detection model or a data augmentation strategy, we focus on understanding the input cloud data and on how to manipulate it so as to improve the performance of existing 3D detection models. Motivated by the non-uniform density distribution problem, we performed experiments (see Fig. \ref{Motivation_random_sampling_bar_plots}) where we reduced the point density of near-by distance ranges of a point cloud and observed that the overall performace on the Centerpoint 3D object detection model ~\cite{yin2021center} improved while also improving the performance at farther away ranges, which suggests the existence of a learning bias toward the closer denser objects due to density imbalance. 
In this paper we propose to manipulate the density of the input point cloud at different distance ranges separately in a pre-processing stage to unlock potential mAP performance improvements of existing 3D object detection models. We propose a model-free input space pre-processing mechanism that partitions the point cloud into Euclidean distance ranges (rings), and performs different levels of density reduction at different ranges. To improve object detection performance we estimate density reduction hyper-parameters using a proposed iterative mechanism based on 
Markov Chain Monte Carlo (MCMC) ~\cite{bishop2006pattern}. We then use density-modified point clouds to train LiDAR 3D object detectors. 


\begin{figure*} 
	\centering	
		\includegraphics[width=1.2\columnwidth, trim={0cm 4cm 5cm 0cm},clip]{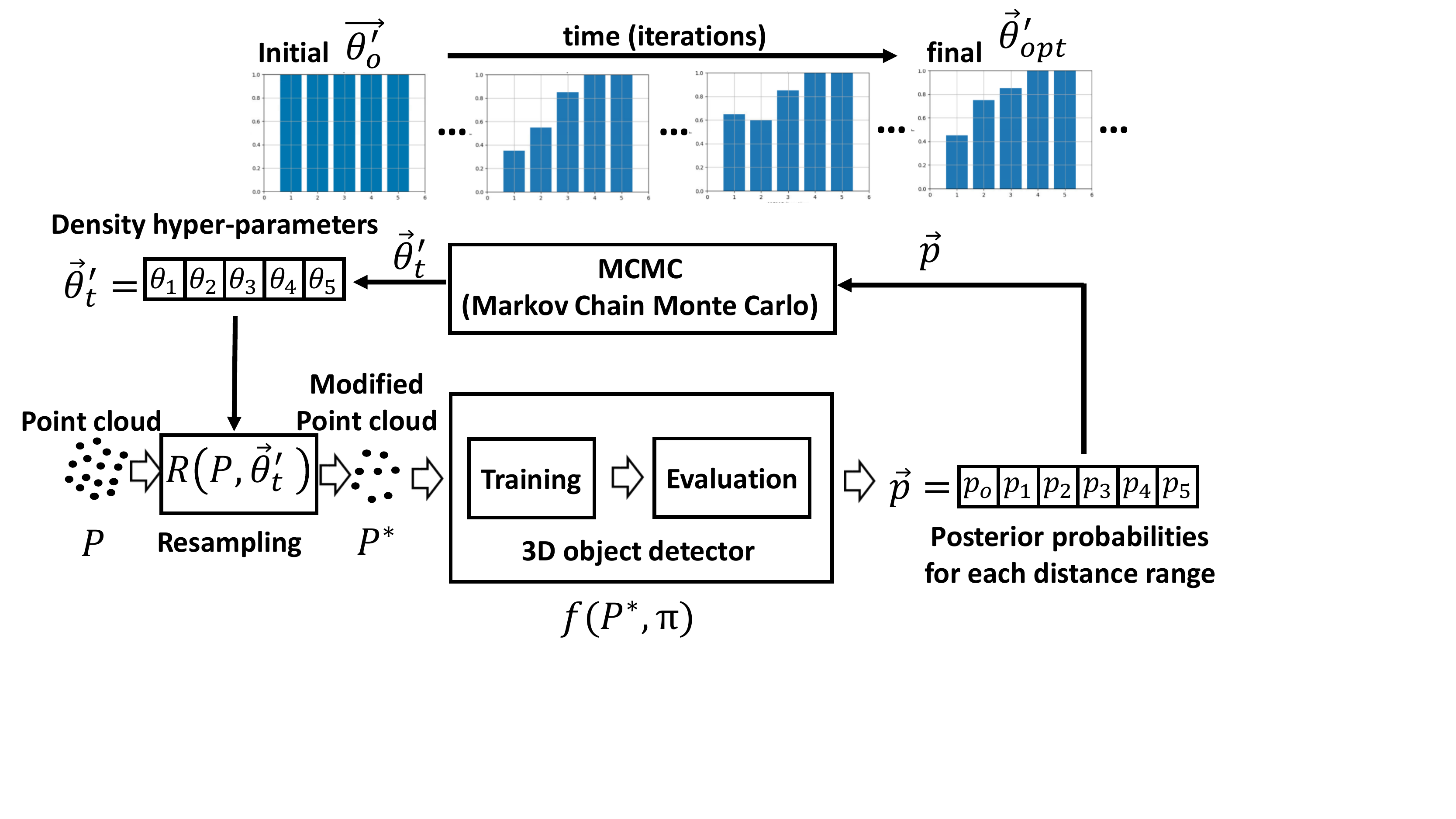} 		
	\caption{LiDAR point cloud per-range density optimization framework. 3D points from different distance ranges are resampled using different parameter settings $\vec{\theta'}$ estimated via iterative MCMC optimization. At each MCMC iteration we initialize the detector $f$ with a fixed pretrained model, and run a one-epoch finetuning on a small subset of data until we obtain the best $\vec{\theta'_{opt}}$ that maximizes 3D detector performance. We then set $R$ to use the fixed $\vec{\theta'_{opt}}$ and run a supervised training of $f$ from scratch to obtain the final results reported in Section \ref{sec:experiments}.
	\label{MCMC_optimization_architecture} }
\end{figure*}


\textbf{Summary of our contributions:}
1) A model-free pre-processing stage that modifies the density of the input raw point cloud (in input space) at different distance ranges. 
2) An MCMC-based iterative mechanism that performs point cloud resampling hyper-parameter optimization via maximizing the performance of LiDAR 3D object detectors.  
3) We perform extensive evaluation on two LiDAR datasets and four LiDAR 3D object detectors where we demonstrate that it is possible to improve the 3D detector performance via our proposed pre-processing stage without architectural changes to the detector.


\section{Method}
\label{sec:method}

Our model-free input space point cloud density manipulation framework is illustrated in Fig. \ref{MCMC_optimization_architecture}. In this framework, a LiDAR point cloud $P$ is input into a model-free point cloud resampling module $R(P, \vec{\theta})$ which modifies the density of $P$ at five distance ranges using resampling hyper-parameters  $\vec{\theta}=[\theta_{1}, \theta_{2}, \theta_{3}, \theta_{4}, \theta_{5}]$. The output from $R(P,\vec{\theta})$ is $P^*$, a version of $P$ with modified density.  We then use $P^*$ to train a 3D object detector $f(P^*, \pi)$, where $\pi$ denotes the model trainable parameters. For each epoch we run an on-line evaluator using the intersection over the union (IoU)-based average precision AP $\in [0,1]$ metrics computed \emph{separately} for each of the five distance ranges. We interpret this set of AP values as a vector of posterior probailities $\vec{p}=[p_{o},p_{1}, p_{2}, p_{3}, p_{4}, p_{5}]$, where $p_{o}$ denotes the overall AP from all distance ranges together. To estimate optimal values for $\vec{\theta}$, we use an MCMC-based optimization procedure that generates proposals $\vec{\theta'}$ that improve 3D object detection performance iteratively.



\subsection{Partitioning the point cloud into distance range clusters}
\label{pointcloud_partitioning}

We work with the publicly-available Waymo and ONCE LiDAR datasets for autonomous driving. Their LiDAR sensor number of beams, vertical (elevation) $\Delta_{\omega}$ and horizontal (azimuth) $\Delta_{\phi}$ angle resolutions are shown in Table \ref{tab:LiDAR_sensor_characteristics}.

\begin{table}
  \centering
  \begin{tabular}{@{}lccc@{}}
    \toprule
    Dataset        & Beams            & Vert. Res.($\Delta_{\omega}$ deg)     & Horz. Res.($\Delta_{\phi}$ deg)\\
    \midrule
    Waymo        & 64 (main)       & $\sim 0.18^{\circ}$                & $\sim 0.22^{\circ}$\\
    ONCE          & 40                  & Var: $\sim 0.3-1.7^{\circ}$     & $\sim 0.2^{\circ}$\\
    \bottomrule
  \end{tabular}
  \caption{Dataset LiDAR sensor resolutions.}
  \label{tab:LiDAR_sensor_characteristics}
\end{table}

\begin{figure} 
	\centering	
		\includegraphics[width=0.8\columnwidth, trim={8cm 7cm 8cm 3cm},clip]{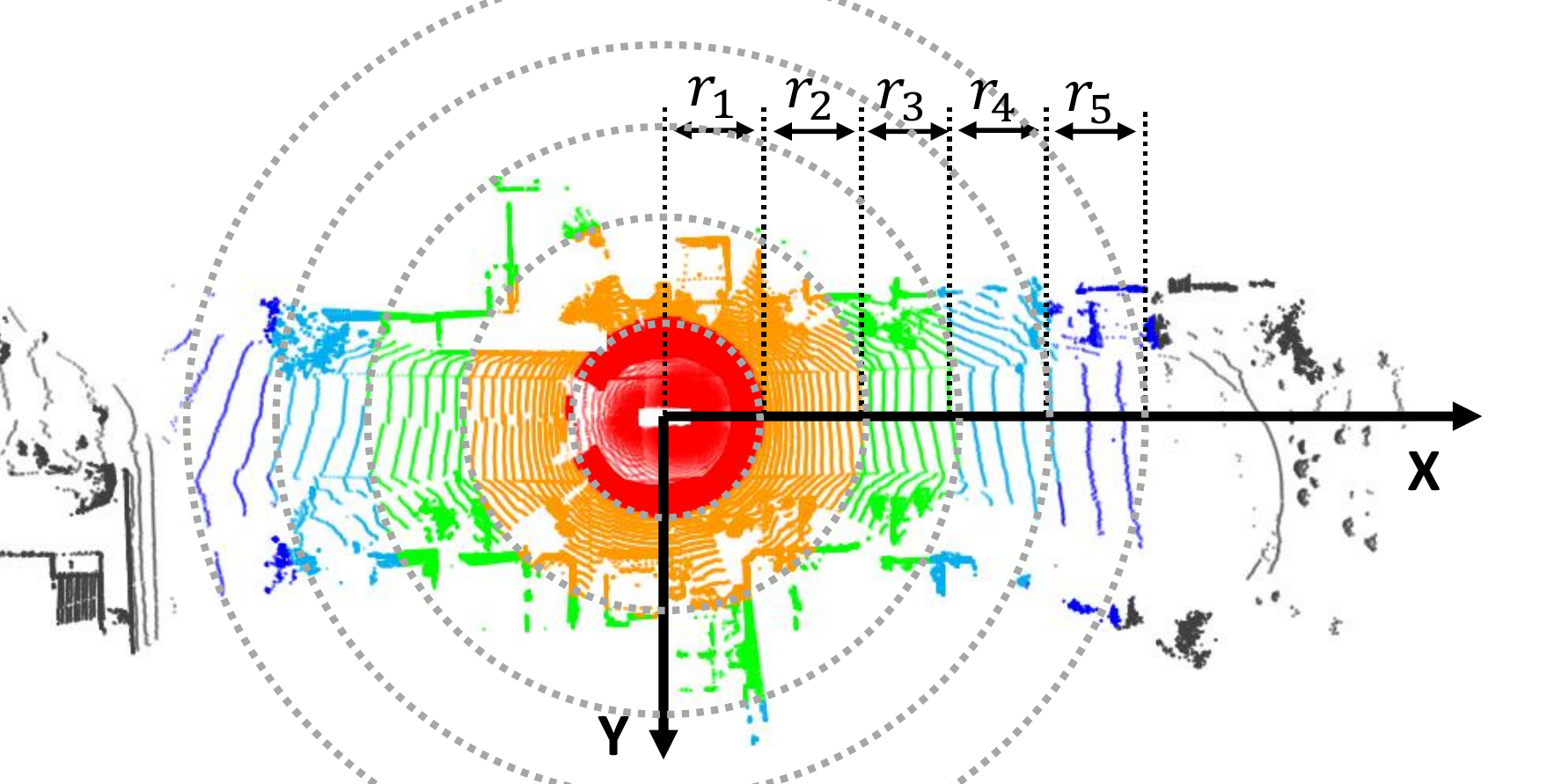} 		
	\caption{Bird's eye view of five Euclidean distance ranges defined on the X-Y plane: $r_1=[0,10$), $r_2=[10,20$), $r_3=[20,30$), $r_4=[30,40$), and $r_5=[40,50$) meters. The colors represent the point cloud clusters $P_1, P_2, P_3, P_4, P_5$.
	\label{Pointcloud_range_clustering} }
\end{figure}

Based on the object-level density distributions from Fig. \ref{Mean_density_bar_plots_2}, we define five non-overlapping 2D Euclidean distance ranges $\vec{r}=[r_{1}, r_{2}, r_{3}, r_{4}, r_{5}]$ on the 3D X-Y plane, where $r_1=[0,10$), $r_2=[10,20$), $r_3=[20,30$), $r_4=[30,40$), and $r_5=[40,50$) meters.  Using these ranges we partition the point cloud $P$ into subsets:  $P = \{ P_1, P_2, P_3, P_4, P_5  \} $, where $P_i  \cap P_j = \emptyset$, $\forall i, j$, as shown in Fig. \ref{Pointcloud_range_clustering}. Each point cloud cluster will be pre-processed separately as explained in the next section.  We leave points located beyond $50$ m untouched (we keep the original points).

\subsection{Point cloud density manipulation}

We implemented two resampling methods $R(P, \vec{\theta})$ to manipulate the density of each $P_i$ 3D point cloud cluster:

\subsubsection{Random sampling}
\begin{figure} 
	\centering	
		\includegraphics[width=1.0\columnwidth, trim={3cm 11cm 3cm 0.5cm},clip]{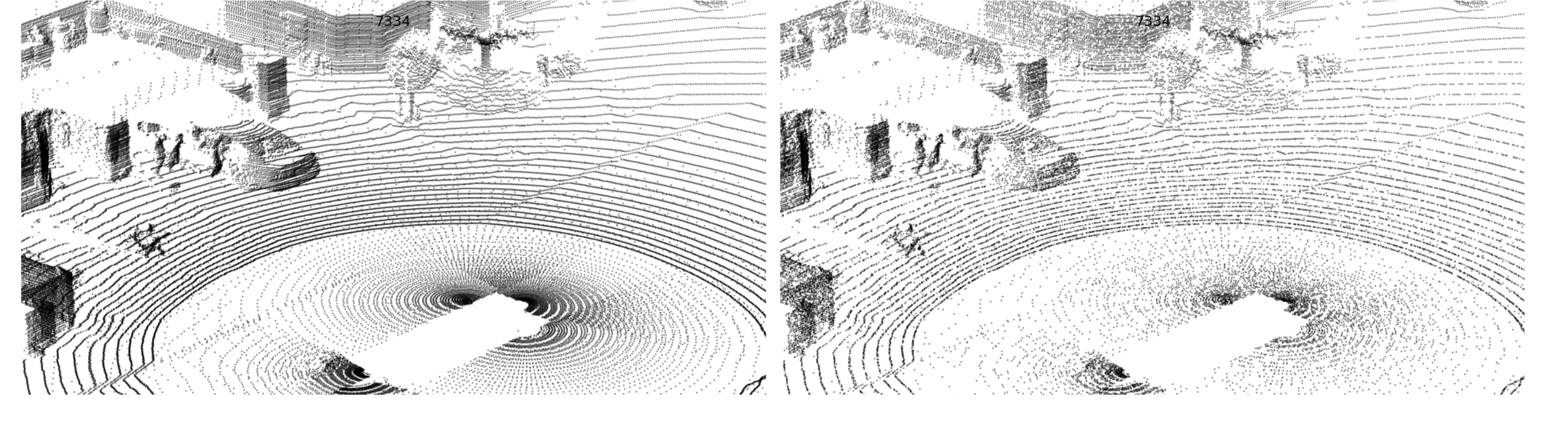} 		
	\caption{Random sampling (RS) on a Waymo point cloud. (left) original point cloud,  (right) clusters $P_1$ and $P_2$ are randomly sampled (keep) by $s_1=0.5$ ($50 \%$) and $s_2=0.75$ ($75 \%$) respectively.
	\label{Pointcloud_random_sampling_example} }
\end{figure}

In the first method, we use random sampling (RS) using a uniform distribution to sample and \emph{keep} a percentage $s_i \in [0,1]$ ($1=100\%$) from the original $N_i$ number of points contained in cluster $P_i$, for each $i \in [1,2,3,4,5]$. 
Fig. \ref{Pointcloud_random_sampling_example} shows an example of RS random sampling on one Waymo point cloud with sampled (keep) percentages $\vec{s}=[0.5, 0.75, 1,1,1]$.  The RS method has an average run time of $0.059$ sec per point cloud, and will be used in the rest of the paper to develop the method and as a baseline in the evaluation section.

\subsubsection{Deterministic grid-based point cloud resampling}

\begin{figure} 
	\centering	
		\includegraphics[width=1.0\columnwidth, trim={1cm 2.75cm 1cm 0cm},clip]{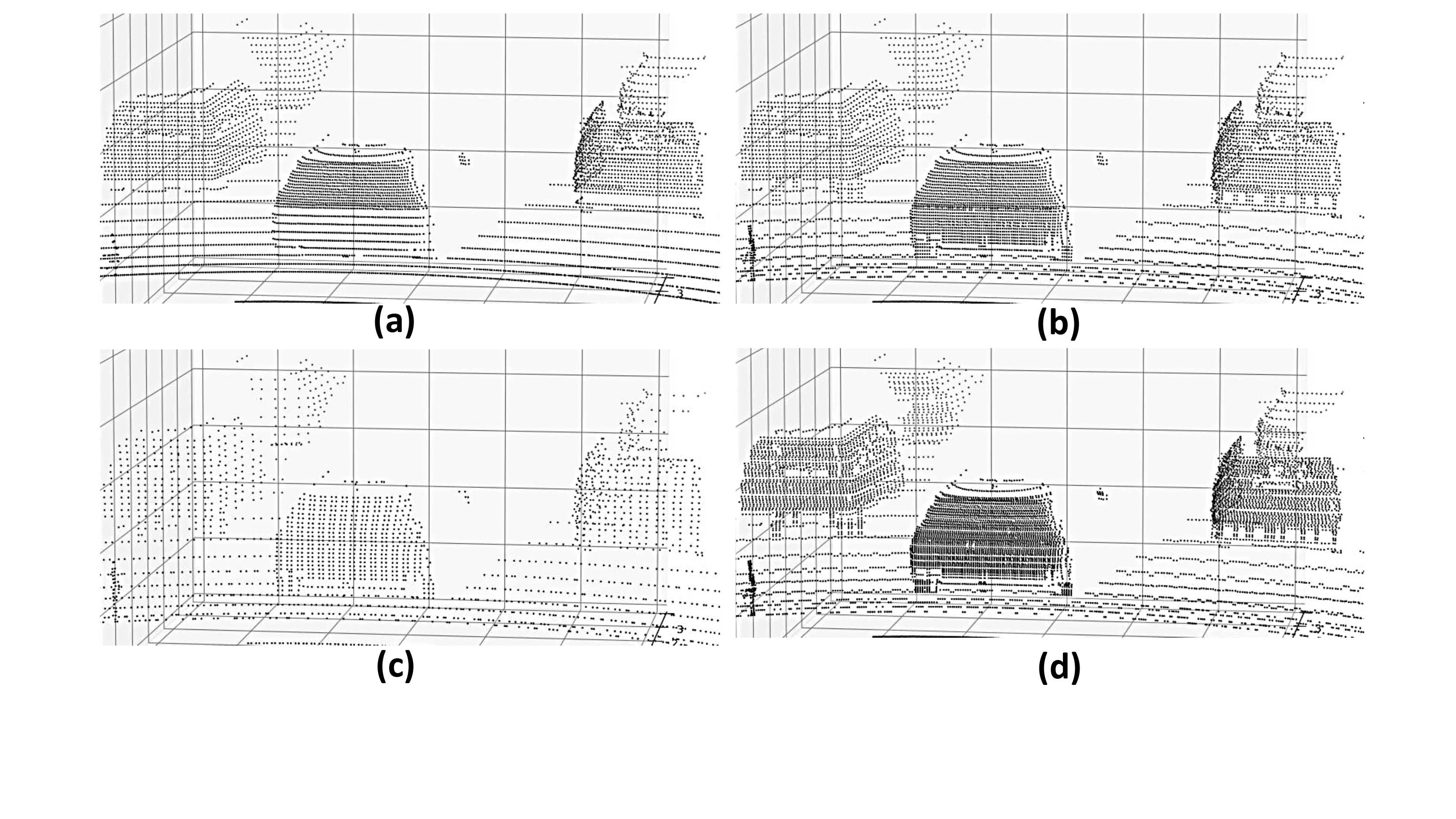} 		
	\caption{Example outputs from the DGR point cloud resampling method on the first distance range $r_1$. (a) Original (ONCE), (b) Interpolating below $-6$ deg (vertical) to $\omega_{d}=0.35$ deg, (c) Resampled to $(\omega_{d}=0.5, \phi_{d}=0.5)$, (d) Resampled to $(\omega_{d}=0.15, \phi_{d}=0.15)$ deg 		     				
	\label{Examples_resampling} }
\end{figure}

We designed and implemented a deterministic grid-based point cloud resampling (DGR) method. DGR works in spherical coordinates ($\omega$=elevation, $\phi$=azimuth), and can either reduce the vertical and/or horizontal resolution of the point cloud, or interpolate points in-between existing points locally using neighbor points. The output point cloud follows the same LiDAR sensor scanning pattern from the input, but resampled to user-defined resolutions. Algorithm \ref{algorithm:DGR} describes the DGR for the resolution reduction case utilized in this paper. The DGR method has an average run time of $0.6$ sec per point cloud. Fig. \ref{Examples_resampling} shows example outputs from the DGR applied to one point cloud from the ONCE dataset.

\begin{algorithm}
\caption{Deterministic grid-based resampling}
\label{algorithm:DGR}
\begin{algorithmic}[1]

\Procedure{DGR}{$P_i$}         
    \State \textbf{Input:} 3D point cloud $P_i$ defined in Sec. \ref{pointcloud_partitioning}
    \State \textbf{Output:} 3D point cloud $P_i^*$ (modified version of $P_i$)
    \State \textbf{Definitions:} 

    \State $\omega$: elevation, $\phi$: azimuth angles of a 3D point (deg.)

    \State $\Delta^{s}_{\omega}$(elev.),$\Delta^{s}_{\phi}$(azimuth) sensor resolutions (deg.) 
    \State $\Delta^{d}_{\omega}$(elev.),$\Delta^{d}_{\phi}$(azimuth) \emph{desired} resolutions (deg.)     
    \State \multiline{ $FOV_{vert}=(\omega_{bot}, \omega_{top})$: LiDAR sensor vertical (azimuth) field of view (deg.) }
    \State \multiline{$T_{norm}=0.25$ Eucl. distance assoc. threshold (m)}
    \State  $N_v= FOV_{vert}/\Delta^{s}_{\omega}$, $N_{dv}= FOV_{vert}/\Delta^{d}_{\omega}$  
    \State  $N_h=360/\Delta^{s}_{\phi}$, $N_{dh}=360/\Delta^{d}_{\phi}$
    \State \textbf{Define grids (a grid is a linear space):}
    \State $G_v=\{\omega:\omega=\omega_{bot}+\Delta^{s}_{\omega}\times n,n \in \{0,1,...N_{v}\}\}$
    \State $G_{dv}=\{\omega:\omega=\omega_{bot}+\Delta^{d}_{\omega}\times n,n \in \{0,1,...N_{dv}\}\}$
    \State $G_h=\{\phi:\phi=0+\Delta^{s}_{\phi}\times n,n \in \{0,1,...N_{h}\}\}$
    \State $G_{dh}=\{\phi:\phi=0+\Delta^{d}_{\phi}\times n,n \in \{0,1,...N_{dh}\}\}$
    \State \multiline{ Split 3D points from $P_i$ into $N_v$ elevation angle clusters $P_i=\{C_1, C_2, ... C_{N_v}\}$ }
    \State  Initialize $P_i^*=\{\}$ 
    \For {$n = 1 \to N_v$}
		\State $pts_{n}=C_n$		
		\State $pts_{w}=\{C_{n+k}\} \forall  k=[-\frac{w}{2}, \frac{w}{2}], k \neq n$,	 
           \State  \multiline{$w$ is a vertical (elevation) neighborhood window}
			    \For {$p_a \in pts_{n}$}
					\State \multiline{Find neighbor points $pts_{b}=\{ p_b \}\in pts_{w}$, such that:$\| p_a\| - \|p_b\| < T_{norm}$}
                           \State \multiline{Compute mean 3D norm $norm_{\mu}$ from $pts_{b}$}
                           \State Compute output 3D point: $p_{out}=p_a$: 
                           \State   \multiline{ Set $p_{out}$ elevation $\omega_{out}$ and azimuth $\phi_{out}$ angles to their nearest location in $G_{dv}$ and $G_{dh}$,}                            
				      \State  \textbf{if} no points in $P_i^*$ have $\omega=\omega_{out},\phi=\phi_{out}$:
                           \State               Scale $p_{out}$ such that $\|p_{out}\|=norm_{\mu}$
                     		\State               Add $p_{out}$ to $P_i^*$
			    \EndFor
    \EndFor
\EndProcedure

\end{algorithmic}
\end{algorithm}

\subsection{Estimating optimal resampling parameters}

Up to this point we have two methods for manipulating the density of the point cloud $P$. In the RS method, the hyper-parameter vector $\vec{s}=[s_1, s_2, s_3, s_4, s_5]$ can be used to control the sampling (keep) percentage of the number of points contained in each cluster $P_i$. On the other hand, $\vec{\Delta_{\omega}}=[\Delta_{\omega_1},\Delta_{\omega_2},\Delta_{\omega_3},\Delta_{\omega_4},\Delta_{\omega_5}]$ can be used to control the DGR resampling angle resolution at each point cloud distance range (we set $\Delta_{\omega}=\Delta_{\phi}$ (isotropic) in the DGR). The question now is what values of $\vec{s}$ and $\vec{\Delta_{\omega}}$ we shall use to obtain performance improvements when training a LiDAR 3D object detector with the modifed point cloud $P^*$?. Optimizing these hyper-parameters manually would be unfeasible and extremely time-consuming, since $\vec{s}, \vec{\Delta_{\omega}} \in \mathbb{R}^{5}$ and the detector would need to be trained and evaluated on $P^*$ for each $\vec{s}$ or $\vec{\Delta_{\omega}}$. Instead, we propose to use approximate inference utilizing Markov Chain Monte Carlo (MCMC) ~\cite{bishop2006pattern}. The MCMC is typically utilized to estimate the parameters of an unknown underlying distribution by generating parameter proposals in an iterative fashion that lead to the maximization of an observed posterior probability. In our problem, we setup the evaluator of the 3D object detector to use the intersection over the union (IoU) -based average precision (AP) metrics. We modified the evaluator to compute both, the overall AP which covers all distance ranges, as well as the AP for each of the five distance ranges from $\vec{r}$ separately. The output of the evaluator is a set of AP values which we interpret as the vector of posterior probailities $\vec{p}=[p_{o}, p_{1}, p_{2}, p_{3}, p_{4}, p_{5}]$, where $p_{o}$ is the overall AP, and $p_i \in [0,1], \forall i$. We then formulate our problem as estimating a set of (approx.) optimal hyper-parameter values $\vec{\theta'_{opt}}$ that maximize the overall posterior probability $p_{o}$.

\subsubsection{MCMC proposals}

In MCMC a proposal distrubution is selected from which to draw samples of the proposed parameters. Inspired by ~\cite{corral2017slot}, in our problem we design the proposal distribution as follows:

\begin{enumerate}
	\item We quantize the hyper-parameter space and draw samples using discrete step perturbations of size $\delta_{\theta}= 0.05$. The sign of the perturbation is drawn from a binomial distribution ~\cite{bishop2006pattern} with $p=0.5$. 

    \item For the Markvovian chain, we allow only one element from $\vec{\theta}$ to be perturbed using $\delta_{\theta}$. In our problem, the point cloud density at closer ranges (e.g. $0$ to $10$ m) is much higher than at ranges farther away (see Fig. \ref{Mean_density_bar_plots_2}). Therefore, the first elements from $\vec{\theta}$ (closer ranges) should be perturbed more and more often than its last elements. We draw the element selection $index$ from a zero-mean 1D Gaussian distribution $j \sim \mathcal{N}( 0,\,\sigma^{2})$, with $\sigma=0.5$, where $index = |round(j)| $, and $index \in [1,5]$. We then use $index$ to select the element from $\vec{\theta}$ that will be perturbed to generate a proposal.  
 
    \item We compute a proposal as :  $\vec{\theta'_{t}}=\vec{\theta'_{t-1}}$, and then, $\vec{\theta'_{t}}[index]=\vec{\theta'_{t}}[index] + \delta_{\theta}$, where the subscript $t$ denotes the current MCMC iteration.
\end{enumerate}

\section{Point cloud resampling and detector training procedure}
\label{sec:training_procedure}

We run our experiments using the procedure described below (the results reported in section \ref{sec:experiments} were obtained following this procedure): 

\begin{enumerate}
	\item We use a dataset $X$ without resampling to perform a supervised training of the detector $f(P, \pi)$ (without MCMC), and save the best model checkpoint $\hat{\pi}$.
 
     \item From $X$, we select (deterministic sampling) subsets $X_{1000}$ and $X_{250}$ of $1000$ training and $250$ validation point clouds respectively. We use $X_{1000}$ and $X_{250}$ in the MCMC iterative procedure.

     \item We then run the MCMC as described in algorithm \ref{algorithm:MCMC_iterations}:

		\begin{algorithm}
		\caption{MCMC iterations}
		\label{algorithm:MCMC_iterations}
		\begin{algorithmic}[1]
                \State Initialize $\vec{\theta'_{o}}=[1,1,1,1,1]$, $\vec{p_o}=[0,0,0,0,0,0]$
                \State $N_{iter}=500$ number of iterations		
			\For {$t = 1 \to N_{iter}$}
				\State Initialize detector $f$ with pretrained weights $\hat{\pi}$    
                  \State Generate proposal $\vec{\theta'_{t}}$ 
				\State \multiline{Using $X_{1000}$,$X_{250}$, run one epoch of detector finetuning/evaluation with $P^*=R(P, \vec{\theta'_{t}})$}
				\State  Accept or reject $\vec{\theta'_{t}}$ (see ~\cite{bishop2006pattern})
                  \State  Update $\vec{\theta'_{opt}}=\vec{\theta'_{t}}$ that maximize $\vec{p_o}$
			\EndFor
		\end{algorithmic}
		\end{algorithm}
		
     \item We use the best $\vec{\theta'_{opt}}$ to start a standard supervised training/evaluation (from scratch) of the 3D detector $f(P^*, \pi)$, with $R(\vec{\theta'_{opt}})$. We discard the 3D detector checkpoints from all MCMC iterations, since we are only interested in obtaining $\vec{\theta'_{opt}}$.
 
\end{enumerate}

\subsection{Training details}
For the supervised pre-training of $f$, we use $4$ GPUs, batch size $32$, and $100$ epochs. For the finetuning of $f$ done during the MCMC procedure, we use $1$ GPU, batch size $1$, and $1$ epoch per iteration. All experiments were run on Nvidia Tesla V100 GPUs.
In all cases, we use Adam optimizer and learning rate set to $0.003$ with one-cycle scheduler with division factor set to $10$.

\begin{table*}
  \centering
  \begin{tabular}{@{}lcccccccc@{}}
    \toprule
                               					& 							 	& 					& 					&  mAP 					&per range			&  					&  					&   \\
    Experiment                           	& MCMC $\vec{\theta'_{opt}}$  	& Overall			& 0-10m 			& 10-20m 		& 20-30m 		& 30-40m 		& 40-50m 		& $>$ 50 m  \\
    \midrule
1 Centerpoint ~\cite{yin2021center}	&                           		& 42.40  			& 61.49 			& 58.34 			& 48.57	 		& 38.45 			& 32.86	 		& 22.00   \\
2 Centerpoint+MCMC-RS                	& [0.75,1,1,1,1]  				& 42.79 			& 62.59 			& 58.87 			& 50.78 & 37.47 			& 34.71	& 22.16  \\
3 Centerpoint+MCMC-DGR (ours)      	& $\omega_1=0.475$     	& \textbf{43.09} & 63.15	& 59.13	& 50.02 			& 39.93	& 33.46 			& 22.09  \\ \hdashline
4 PV-RCNN ~\cite{shi2020pv}         	&                            		& 40.43			 & 56.48			    & 53.26			& 45.36			& 39.01 & 31.96			& 22.72  \\
5 PV-RCNN+MCMC-RS                    	& [0.55,0.9,0.9,1,1]    		& 41.49                & 58.16	& 54.85 & 46.16 & 38.78 		& 33.30 			& 23.95   \\
6 PV-RCNN+MCMC-DGR (ours)      		& $\omega_1=0.55$        & \textbf{41.92}	    		 & 64.60  			& 54.45    	    	& 46.87    	  	& 39.22    		& 33.89 & 23.17          \\ \hdashline
7 PDV ~\cite{hu2022point}             	&          						& 40.81 			& 60.10  			& 52.71   		 & 45.17 			     & 37.21 		 	 & 32.47 			    & 24.27  \\
8 PDV+MCMC-RS           					& [0.65,0.9,1,1]    			& 41.33 			& 60.38 & 52.47 			 & 46.76 	& 38.16 			 & 34.03	& 23.87  \\ 
9 PDV+MCMC-DGR (ours)           		& $\omega_1=0.525$     	& \textbf{41.41} & 58.42  & 55.07 & 45.18  			& 39.09 & 33.81     		    & 22.86   \\ \hdashline
10 PointRCNN ~\cite{shi2019pointrcnn}   	&          					& 33.84 			& 59.68  			& 52.17   		 & 38.53 			     & 29.22 		 	 & 23.19 			    & 12.53  \\
11 PointRCNN+MCMC-RS           			& [0.65,0.9,1,1]    		& 36.62 			& 59.45 & 52.85 			 & 41.77 	& 33.07 			 & 26.99	& 15.67  \\ 
12 PointRCNN+MCMC-DGR (ours)         	& $\omega_1=0.525$    	& \textbf{37.35} & 58.51  & 54.16 & 42.9  			& 32.91 & 27.76     		    & 15.99   \\ 
    \bottomrule
  \end{tabular}
  \caption{Point cloud resampling detector experiments with ONCE dataset. To observe the effects of the point cloud density manipulation on 3D detection performance, we disabled all augmentations.}
  \label{tab:once_results}
\end{table*}

\begin{table*}
  \centering
  \begin{tabular}{@{}lcccccccc@{}}
    \toprule
                               					& 							 	& 					& 					&  mAP 					&per range			&  					&  					&   \\
    Experiment                          		& MCMC $\vec{\theta'_{opt}}$	& Overall			& 0-10m 			& 10-20m 		& 20-30m 		& 30-40m 		& 40-50m 		& $>$ 50 m  \\
    \midrule
1 Centerpoint ~\cite{yin2021center}   	&                           		& 41.32			& 71.67			& 67.36			& 57.37			& 45.62	& 30.27			& 18.12 \\      
2 Centerpoint+MCMC-RS  				& [0.65,0.75,1,1,1]			& 41.94			& 72.54	& 67.79			& 57.53	& 44.9	    			& 30.15			& 19.00 \\         	
3 Centerpoint+MCMC-DGR (ours)  		& $\omega_1=0.50$       	& \textbf{42.48}	& 71.64			& 69.60	& 57.16			& 44.71			& 32.19	& 18.40 \\ \hdashline    
4 PV-RCNN ~\cite{shi2020pv}  			&                            		& 37.35			& 63.81			& 59.35			& 48.97			& 40.11 & 28.94	& 18.47 \\    
5 PV-RCNN+MCMC-RS                     	& [0.35,0.85,1,1,1]    		& 37.48 			& 66.99 & 59.02   		& 50.14  & 38.15    		& 27.47   			& 18.99  \\   	  
6 PV-RCNN+MCMC-DGR (ours)           	& $\omega_1=0.55$      & \textbf{37.91}	& 64.47    		& 60.39 & 49.88    		  & 38.5	     	& 28.82  			    & 19.36       \\ \hdashline  	  
7 PDV ~\cite{hu2022point}    			&          						& 38.49			& 68.48   			& 62.15   			& 50.32 			& 42.8	& 28.76			& 16.94 \\
8 PDV+MCMC-RS                            	& [0.55,1,1,1,1]  				& 38.47   			& 66.48   			& 63.36	& 48.38     		& 42.06   			& 29.59	& 17.07         	   \\
9 PDV+MCMC-DGR (ours)                 	& $\omega_1=0.30$       	& \textbf{39.09} & 69.58	& 60.48   			& 52.83	& 41.13   			& 29.01    		& 18.63	  \\ \hdashline
10 PointRCNN ~\cite{shi2019pointrcnn}   	&          					& 25.43 			& 56.38  			& 49.98   		 & 33.94 			     & 15.57 		 	 & 19.33 			    & 10.58  \\
11 PointRCNN+MCMC-RS           			&[0.55,1,1,1,1]    		& 26.33 			& 57.88 & 52.85 			 & 34.33 	& 19.26 			 & 19.67	& 9.85  \\ 
12 PointRCNN+MCMC-DGR (ours)         	& $\omega_1=0.30$    	& \textbf{26.95} & 50.72  & 51.84 & 38.23  			& 21.82 & 18.48     		    & 9.85   \\ 
    \bottomrule
  \end{tabular}
  \caption{Point cloud resampling detector experiments with Waymo dataset. Again, we disabled all augmentations to be able to observe the effects of the point cloud density manipulation on 3D detection performance.}
  \label{tab:waymo_results}
\end{table*}

\section{Experiments and results}
\label{sec:experiments}

To focus on (and be able to see) the effects of the point cloud density manipulation using the RS and DGR methods on 3D detection performance, all our experiments are performed without data augmentations (no rotations/translations/flip/scalings, and no database object insertion). 


\subsection{Experiments with different 3D object detectors}
\label{sec:experiments_with_detectors}
We conduct our main experiments with four state-of-the-art LiDAR 3D object detectors, namely, Centerpoint ~\cite{yin2021center}, PV-RCNN ~\cite{shi2020pv}, PDV ~\cite{hu2022point}, and PointRCNN ~\cite{shi2019pointrcnn}. We use both the Waymo and ONCE datasets to train and evaluate these detectors.  Centerpoints performs a height compression step, and we want to observe how it reacts when points are resampled in 3D space. Both, PV-RCNN and PDV use a mixture of voxelization (no height compression), and point-level feature learning, while PDV takes point cloud density into consideration. PointRCNN is a point-based method. Our goal is then to understand the effects of resampling the point cloud on the performance of these detectors. With this goal in mind, the number of point clouds we use for ONCE is $4961$ for training and $3311$ for evaluation, which correspond to the original dataset size. For Waymo we use a $5\%$ dataset split with $7905$ point clouds for training and $2000$ for evaluation. These splits are chosen deterministically using dataset sampling parameters available in the OpenPCDet ~\cite{openpcdet2020} base code that we use in this paper. To avoid introducing domain shifts ~\cite{wang2020train, yang2021st3d, corral2021lidar} between the dataset splits due to resampling of the point clouds at the input, we apply the resampling $R(P, \vec{\theta})$ to \emph{both}, the training and validation splits. 
We train and evaluate each detector on all of the following object classes:  Waymo: $\{vehicle, pedestrian, cyclist\}$,  ONCE: $\{car, bus, truck, pedestrian, cyclist\}$. Average precision (AP) is computed for each object class using Intersection of the Union (IoU) metric, with threhsolds $0.7$ (for all vehicles), $0.3$ for pedesrians, and $0.5$ for cyclists. We report the mean AP (mAP) for each of the five distance ranges, as well as overall (which covers all ranges). The results are shown on Table  \ref{tab:once_results} for ONCE, and on  Table \ref{tab:waymo_results} for Waymo$5\%$

\subsection{Results discussion}
\begin{figure} 
	\centering	
		\includegraphics[width=1.0\columnwidth, trim={2cm 3.5cm 2cm 0.5cm},clip]{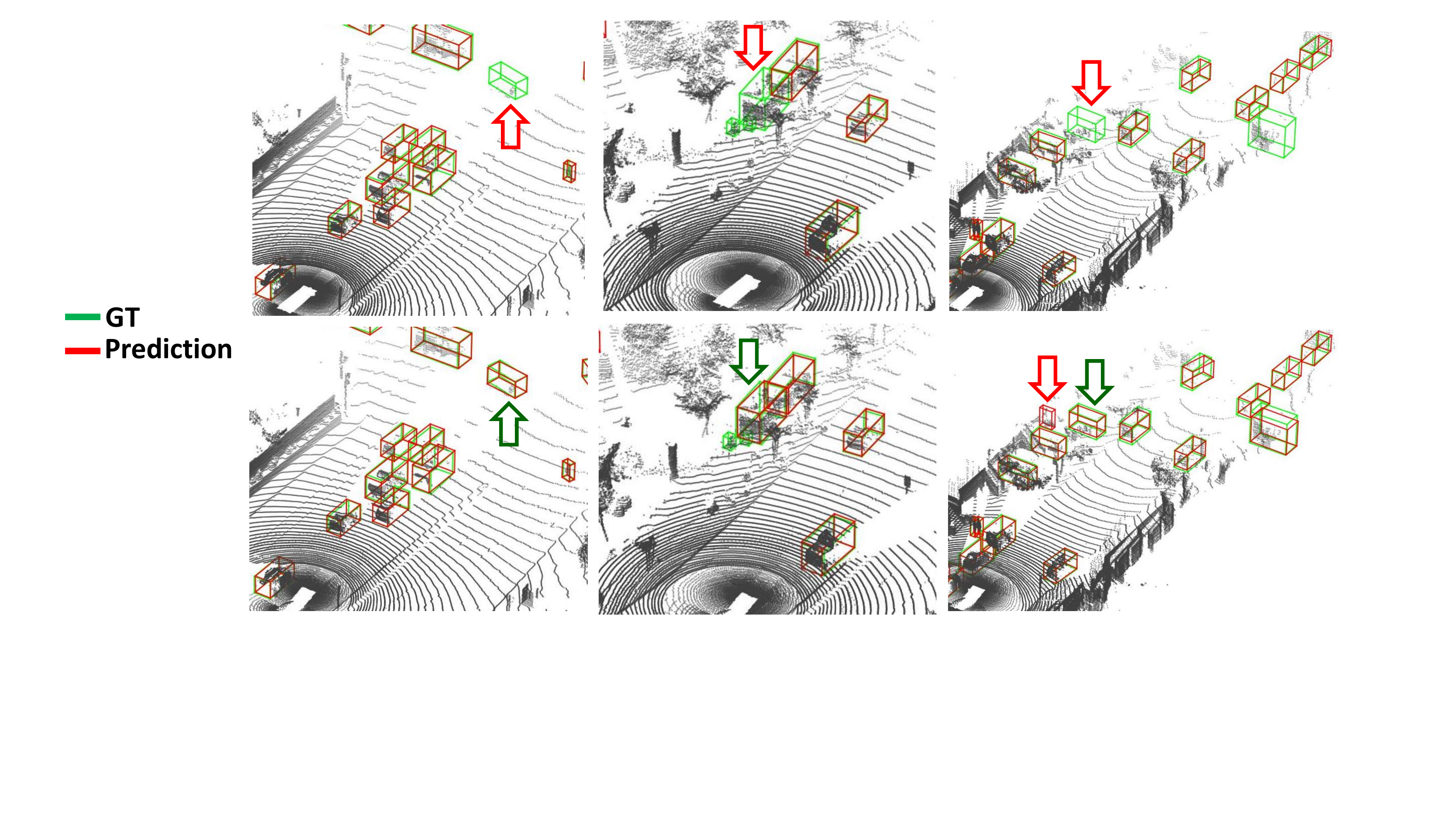} 		
	\caption{Qualitative outputs. Top row: outputs with detector trained with original data (see false negatives). Bottom row: detector trained with DGR data. Farther objects are now detected. 		     				
	\label{Qualitative_examples} }
\end{figure}

In LiDAR 3D object detection, the performance improvements contributed by emerging state-of-the-art detectors is typically small, in the order of $1\%$ or less (See ~\cite{geiger2013vision} benchmark for example). Our goal is to squeeze out any performance improvement from manipulating the density of the input point cloud. From the results from Tables \ref{tab:once_results} and \ref{tab:waymo_results} we make the following observations: 1) \emph{On average} (over detectors), training the detectors with data processed with the GDR improves the mAP (compared to training with the original data) by $1.57$ on ONCE, and by $0.96$ on Waymo. The maximum observed GDR improvement of $3.51$ is seen on PointRCNN with ONCE (Table \ref{tab:once_results} rows $10$ and $12$). 2) In general, the overall results obtained with the GDR outperform those from the RS and baseline. This is also the case for the first two distance ranges ($r_1$, and $r_2$). 3) On both Waymo and ONCE datasets, Centerpoint outperforms both PV-RCNN (consistent with ~\cite{yin2021center}) and PDV. However, PDV seems to outperform PV-RCNN mainly on Waymo, which is consistent with the results reported by the authors of ~\cite{hu2022point}. 4) DGR helped PDV to outpeform its own baseline  (Table \ref{tab:waymo_results} rows $7$ and $9$), which suggests that manipulating the point cloud in input space on top of the PDV voxel feature-level density processing can contribute jointly to the improved detection performance. Fig. \ref{Qualitative_examples} shows some qualitative detector outputs.

\subsection{Ablation study}
\label{sec:ablation}

To better understand the effects of resampling the point cloud with different settings, we perform the following ablation study with the Centerpoint 3D object detector, on the ONCE dataset. We resample the \emph{whole} point cloud using a \emph{single} distance range $r=[0, d)$, while varying its upper limit $d$. For these experiments we use DGR. 

\textbf{1. Varying the distance range upper limit $d$}. Table \ref{tab:ablation_once_varying_range} rows $1-4$ show the effects of reducing the distance range upper limit $d$ from $75$ m down to $10$ m using \emph{fixed} default grid resampling settings $(\Delta_{\omega}=0.35 deg, \Delta_{\phi}=0.2 deg)$ selected based on Table \ref{tab:LiDAR_sensor_characteristics}. We observe that resampling the point cloud at far away ranges affects the detector performance negatively, since those far regions of the point cloud are originally sparse, and should not be processed. 

\textbf{2. Varying the grid resampling resolution angle $\Delta_{\omega}$}.
We now fix the range distance upper limit $d=10$m, and start varying the angle resolution $\Delta_{\omega}$ of the grid resampling stage, starting from $0.35$ deg (denser), up to $1.0$ deg (sparser). Note that we set $\Delta_{\phi}=\Delta_{\omega}$ in the DGR module. The results are shown in rows $5-8$ from Table \ref{tab:ablation_once_varying_range}. From these experiments we observe that there exists an optimal (maxima) set of resampling parameters that maximizes detector performance, which in this case is at around $\Delta_{\omega}=0.5$ deg, after which, increasing the value of $\Delta_{\omega}$ further (which in turn reduces density), degrades the performance because smaller objects such as pedestrians start becoming too sparse.  Therefore, there is a trade-off between reducing the density of the point cloud and the size of objects that need to be detected. 

\textbf{Interpolation}.
In our last experiment (row $9$ from Table \ref{tab:ablation_once_varying_range}), we kept the settings from row $6$ from the table, but enabled 3D point interpolation by setting the angle to $\Delta_{\omega}=0.15$ deg for the $10$-$30$ m range. Although interpolation may produce appealing point cloud visualizations (see Fig. \ref{Examples_resampling}(d)), it did not help boost performance. In fact peformance was reduced slighty, while slowing down our experiments since the size of the point cloud increased due to the interpolation. For this reason we mainly focused on reducing density (while avoiding interpolation) in section \ref{sec:experiments_with_detectors}.


\begin{table}
  \centering
  \begin{tabular}{@{}lcccccc@{}}
    \toprule
                          & $(\Delta_{\omega}, \Delta_{\phi})$   & Range   & Overall  & Veh. & Ped.  & Cyclist  \\
    \midrule
    1                    & $(0.35, 0.20) $    & 0-75m & 36.92	   & 45.69	   & 27.04	        & 38.01    \\
    2                    & $(0.35, 0.20) $    & 0-30m & 40.72	   & 48.19	   & 31.10	        & 42.87    \\
    3                    & $(0.35, 0.20) $    & 0-20m & 42.16	   & 49.86	   & 31.92	        & 44.72    \\
    4                    & $(0.35, 0.20) $    & 0-10m & 42.63	   & 49.87	   & 32.68	        & 45.34    \\ \hline 
    5                    & $(0.35, 0.35) $    & 0-10m & 42.95	   & 49.57	   & 33.83	        & 45.47    \\
    6                    & $(0.50, 0.50) $    & 0-10m & \textbf{43.09}	   & 49.86	   & 33.84	        & 45.57    \\
    7                    & $(0.75, 0.75) $    & 0-10m & 42.84	   & 49.93	   & 33.48	        & 45.12    \\
    8                    & $(1.00, 1.00) $    & 0-10m & 42.66	   & 49.80	   & 32.84	        & 45.34    \\ \hline 
    9                    & $(0.15, 0.15) $    & 10-30m & 42.98	   & 49.82	   & 33.59	        & 45.52    \\

    \bottomrule
  \end{tabular}
  \caption{Ablation experiments with Centerpoint with ONCE dataset.  (rows 1-4): Fixing the DGR settings and varying the distance range max. limit, (rows 5-8): Fixing the distance range max. limit, and varying the DGR resolution angle $\Delta_{\omega}$, (row 9): DGR interpolation for range 10-30m.}
  \label{tab:ablation_once_varying_range}
\end{table}


\section{Conclusions}
\label{sec:conclusions}


In this paper we have shown that our proposed framework for input space density manipulation of the 3D point cloud at different distance ranges can help improve the performance of different types of 3D object detectors. We presented two methods to manipulate the point cloud density, and proposed a methodology for optimizing their hyper-parameters for detection performance by means of an iterative procedure based on MCMC. We demonstrated the effectiveness of our framework on four state-of-the-art LiDAR 3D detectors on two widely-used public datasets. We hope that our results can inspire new designs of LiDAR 3D object detectors that improve the handling of the density imbalance problem.

\clearpage

\bibliographystyle{IEEEtran}
\bibliography{egbib_density_iros_2023}

\end{document}